\pdfoutput=1

\documentclass[11pt]{article}

\usepackage{emnlp2021}

\usepackage{times}
\usepackage{latexsym}
\usepackage{graphicx}
\usepackage{tabularx}
\usepackage{subcaption}
\usepackage{soul}
\usepackage{hyperref}
\usepackage{url}
\usepackage{booktabs}
\usepackage[T2A,T1]{fontenc}
\usepackage[utf8]{inputenc}
\usepackage[russian,english]{babel}
\usepackage{todonotes}
\usepackage[normalem]{ulem}
\usepackage{float}
\useunder{\uline}{\ul}{}

\usepackage{microtype}
\makeatletter
\newcommand\footnoteref[1]{\protected@xdef\@thefnmark{\ref{#1}}\@footnotemark}
\makeatother
\title{Pushing the Right Buttons: Adversarial Evaluation of Quality Estimation}

\author{
Diptesh Kanojia$^{1}$,
Marina Fomicheva$^{2}$, 
Tharindu Ranasinghe$^{3}$,\\
\textbf{Fr\'ed\'eric Blain}$^{4}$,
\textbf{Constantin Or\u asan}$^{5}$,
\textbf{Lucia Specia}$^{6}$\\
  $^{1,5}$Centre for Translation Studies, University of Surrey \\
  $^{2}$University of Sheffield
  $^{3,4}$University of Wolverhampton
  $^{6}$Imperial College London\\
  $^{1,5}$\texttt{\{d.kanojia,c.orasan\}@surrey.ac.uk}, 
  $^{2}$\texttt{m.fomicheva@sheffield.ac.uk},\\
  $^{3,4}$\texttt{\{t.d.ranasinghehettiarachchige,f.blain\}@wlv.ac.uk},\\
  $^{6}$\texttt{l.specia@imperial.ac.uk}
}

\begin{document}
\maketitle
\begin{abstract}
Current Machine Translation (MT) systems achieve very good results on a growing variety of language pairs and datasets. However, they are known to produce fluent translation outputs that can contain important meaning errors, thus undermining their reliability in practice. Quality Estimation (QE) is the task of automatically assessing the performance of MT systems at test time. Thus, in order to be useful, QE systems should be able to detect such errors. However, this ability is yet to be tested in the current evaluation practices, where QE systems are assessed only in terms of their correlation with human judgements. In this work, we bridge this gap by proposing a general methodology for adversarial testing of QE for MT. First, we show that despite a high correlation with human judgements achieved by the recent SOTA, certain types of meaning errors are still problematic for QE to detect. Second, we show that on average, the ability of a given model to discriminate between meaning-preserving and meaning-altering perturbations is predictive of its overall performance, thus potentially allowing for comparing QE systems without relying on manual quality annotation.
\end{abstract}

\section{Introduction}
\label{sec:intro}

Quality Estimation (QE) is the task of predicting the quality of Machine Translation (MT) output in the absence of human reference translation. Recent QE models based on multilingual pre-trained representations~\citep{ranasinghe-etal-2020-transquest-translation} have shown impressive results achieving up to 0.9 Pearson correlation with human judgements of translation quality at sentence level~\citep{specia-etal-2020-findings-wmt}. Not unlike other NLP systems, QE systems are typically tested on held-out datasets. On the one hand, such evaluation usually requires collecting additional human judgements and thus cannot be easily extrapolated to a different usage scenario, for example, a new language pair. On the other hand, evaluation on a given test set can hide performance issues related to the phenomena that are underrepresented in the data but are critical to the reliable performance of the system. Finally, a single  statistic capturing overall performance does not provide any insights on the strengths and weaknesses of a given approach. As a way to overcome these limitations, we explore adversarial evaluation for QE. Specifically, we introduce two types of changes to high-quality MT outputs: meaning-preserving perturbations (MPPs) and meaning-altering perturbations (MAPs). Intuitively, we expect a strong QE system to assign lower scores to the sentences containing MAPs compared to the sentences with MPPs. Based on this intuition, we devise experiments to systematically test a set of five different QE systems by comparing the scores they produce for sentences containing MPPs and MAPs. We use the difference in the predicted scores as a way of detecting specific problems as well as for assessing the overall performance of the systems. Our main findings\footnote{Code available from \url{https://github.com/dipteshkanojia/qe-evaluation}.}
can be summarised as follows:
\begin{itemize}
    \item Overall, SOTA QE models are robust to MPPs and are sensitive to MAPs, thus supporting the claims that such models are indeed strong predictors of MT quality.
    \item SOTA QE models fail to properly detect certain types of MAPs, such as negation omission, which highlights the weaknesses of these models that cannot be detected using standard evaluation methods.
    \item The overall results of our probing experiments on a set of QE models are consistent with their correlation with human judgements. This suggests that the proposed evaluation methodology can be used to assess the performance of QE models with no need for collecting gold standard human annotation.
\end{itemize}

In the remainder of this paper, we first discuss related work on probing for NLP (Section~\ref{sec:rw}). We then describe the dataset (Section~\ref{sec:dataset}) and QE models used in our experiments (Section~\ref{sec:qemodels}). We introduce our probing setup and strategies in Section~\ref{sec:probing} and present and discuss the results in Section~\ref{sec:results}.

\section{Related Work}
\label{sec:rw}

Very few studies have analysed the performance of QE models beyond correlation with human judgements on held-out datasets. To the best of our knowledge, the only work that analyses the behaviour of QE models is~\citet{sun-etal-2020-estimating}. On various datasets popularly used for training QE models, they show that they contain certain biases, such as a skew towards high-quality MT outputs and lexical artefacts that are picked up by the SOTA architectures, \textit{e.g.,} sentences with certain tokens tend to have high or low scores. They also show that QE models can perform very well on these datasets by encoding only the source or target sentences. By contrast, we study the behaviour of the models under specific linguistic conditions. Our experiments show that the models are not sensitive to certain meaning errors, which is in line with~\citep{sun-etal-2020-estimating}'s assumption that SOTA QE models do not capture adequacy.   

For MT, various studies have shown that models can achieve high performance on clean data, they are very brittle to noisy inputs, where both synthetic (e.g. character flips) or natural (social media data) noise is used to probe models~\citep{belinkov2018synthetic,khayrallah-koehn-2018-impact,li-etal-2019-findings,passban2020revisiting}. For other NLP tasks, black-box methods for adversarial evaluation have been proposed that apply meaning-preserving perturbations in order to test whether the models are sensitive to changes in the input~\citep{ribeiro-etal-2018-semantically}. Different from this line of work, we probe the robustness of QE models to spurious changes but also sensitivity to relevant changes, such as meaning errors.~\citet{ribeiro2020beyond} recently devised a general methodology for behavioural testing of NLP models. They generate a subset of simple examples meant to test general linguistic capabilities expected from an NLP system. However, the linguistic capabilities tested within this framework are not directly applicable to the QE task. They could not, for example, capture the ability of a QE system to detect omission errors or copy errors in translation.

\section{Dataset}
\label{sec:dataset}

The dataset used in this paper is a subset from the WMT 2020 Quality Estimation Shared Task 1, sentence-level prediction~\citep{specia-etal-2020-findings-wmt}. This data consists of seven language pairs which can be classified as high-resource [English-German (En-De), English-Chinese (En-Zh)], medium-resource [Russian-English (Ru-En), Romanian-English (Ro-En), Estonian-English (Et-En)], and low-resource [Sinhala-English (Si-En), Nepalese-English (Ne-En)] pairs. Except for Ru-En,  sentences are extracted solely from Wikipedia. The Ru-En data also contains additional sentences from  Reddit~\citep{fomicheva-etal-2020-unsupervised}. The data was collected by machine translating sentences sampled from source-language articles using SOTA NMT models built using the \textit{fairseq} toolkit~\citep{ott-etal-2019-fairseq}. The data was annotated with a variant of Direct Assessment (DA) scores~\citep{graham_baldwin_moffat_zobel_2017} by professional translators. Each translation was rated with a score in 1-100, according to the perceived translation quality by at least three translators~\citep{specia-etal-2020-findings-wmt}. The goal of QE systems built on this data is to predict a \textit{z-score} normalised mean DA for each \textit{test} source-target pairs, which we further standardise between 0 and 1.

In the original dataset, 9K sentences per language pair were randomly split in training (7K), validation (1K) and test (1K). In this study, we focus on probing the models by modifying the target side (translations) with various perturbations. To keep the experiments consistent across the language pairs, we only consider the five pairs with English as the target language. 

We use the standard training partition of the data to train our QE models. To evaluate our probes, the assumption made is that sentences with perturbations should lead to lower predicted QE scores than original sentences. However, this assumption only holds if we can ensure that the original sentences have high enough quality since perturbing very low-quality sentences with already very low scores would not necessarily lead to further degradations. Therefore, we create a subset of the validation + test sets by applying the threshold of 0.7 on the standardised human (DA) scores to reflect high quality, based on the definition of the DA scores used as guidelines for annotators in this dataset. 
Table~\ref{tab:dataset} shows the resulting number of validation + test instances for each language. We hereafter refer to this set as our {\bf test set}.  


\begin{table}[!ht]
\centering
\resizebox{\columnwidth}{!}{%
\begin{tabular}{@{}cccccc@{}}
\toprule
\textbf{Language Pair} & \textbf{Ru-En} & \textbf{Ro-En} & \textbf{Et-En} & \textbf{Si-En} & \textbf{Ne-En} \\ \midrule
\#sentences & 1245 & 1035 & 766 & 404 & 100 \\
Low-resource & No & No & No & Yes & Yes \\ \bottomrule
\end{tabular}%
}
\caption{The number of selected sentences in our test set for each language pair. These are sentences judged to have high-enough quality by human translators. 
}
\label{tab:dataset}
\end{table}
\vspace{-0.5cm}
\begin{table*}[ht!]
\centering
\resizebox{\textwidth}{!}{%
\begin{tabular}{clc}
\hline
\textbf{Source} & \multicolumn{2}{l}{\textbf{În alegerile europarlamentare din 2014, UKIP, partid de extremā dreaptā, a obținut peste 20 de locuri in parlamentul european.}} \\ \hline
\textbf{Reference} & \textbf{In the 2014 European Parliamentary elections, UKIP, a right-wing party, obtained more than 20 seats in the European Parliament.} & \textbf{S1} \\ \hline
\textbf{Translation} & \textbf{In the 2014 European Parliamentary elections, UKIP, party of extreā dreaptā, obtained more than 20 seats in the European Parliament.} & 0.81 \\ \hline
MPP1 & In the 2014 European Parliamentary elections UKIP party of extreā dreaptā obtained more than 20 seats in the European Parliament & 0.79 \\
MPP2 & In the 2014 European Parliamentary elections\textbf{!} UKIP\textbf{(} party of extreā dreaptā. obtained more than 20 seats in the European Parliament\textbf{?} & 0.69 \\
MPP3 & In 2014 European Parliamentary elections, UKIP, party of extreā dreaptā, obtained more than 20 seats in European Parliament. & 0.80 \\
MPP4 & In \textbf{such} 2014 European Parliamentary elections , UKIP , party of extreā dreaptā , obtained more than 20 seats in \textbf{those} European Parliament. & 0.69 \\
MPP5 & \begin{tabular}[c]{@{}l@{}}\textbf{IN} the 2014 \textbf{EUROPEAN} Parliamentary \textbf{ELECTIONS}, UKIP, party of extreā \textbf{DREAPTĀ}, \\ \textbf{OBTAINED} more \textbf{THAN 20 SEATS} in \textbf{THE EUROPEAN PARLIAMENT}.\end{tabular} & 0.76 \\
MPP6 & in the 2014 European parliamentary elections, \textbf{ukip}, party of extreā dreaptā, obtained more than 20 seats in the European Parliament. & 0.75 \\ \hline
\end{tabular}%
}
\caption{An example of each MPP from our dataset for Ro-En. `Translation'  is the original machine translated sentence for the given source sentence, which was assigned an average DA score of 0.70 by human annotators (in 0-1). S1 are scores from the MonoTransQuest architecture. The reference translation is only shown for readability, as it was not used by humans nor QE models.}
\label{tab:mppexamples}
\end{table*}

\begin{table*}[ht]
\centering
\resizebox{\textwidth}{!}{%
\begin{tabular}{@{}clc@{}}
\toprule
\textbf{Source} & \multicolumn{2}{l}{\begin{tabular}[c]{@{}l@{}}\foreignlanguage{russian}{На слушании в декабре Блэквуд сказал, что не имел намерения оскорбить буддизм, когда размещал изображение, а} \\ \foreignlanguage{russian}{после того, как осознал, что оно вызвало массовое возмущение, удалил его и опубликовал извинение.}\end{tabular}} \\ \midrule
\textbf{Reference} & \textbf{\begin{tabular}[c]{@{}l@{}}At a hearing in December, Blackwood said he had not intended to offend Buddhism when he posted the image, and \\ after realizing it had caused widespread outrage, deleted it and issued an apology.\end{tabular}} & \textbf{S1} \\ \midrule
\textbf{Translation} & \textbf{\begin{tabular}[c]{@{}l@{}}At a hearing in December, Blackwood said he had not intended to offend Buddhism when he posted the image, and \\ after realizing it had caused widespread outrage, deleted it and issued an apology.\end{tabular}} & 0.83 \\ \midrule
MAP1 & \begin{tabular}[c]{@{}l@{}}At a hearing in December, Blackwood said he \textbf{had intended} to offend Buddhism when he posted the image, and \\ after realizing it had caused widespread outrage, deleted it and issued an apology.\end{tabular} & 0.82 \\
MAP2 & \begin{tabular}[c]{@{}l@{}}At a hearing \textbf{in,} Blackwood said he had not intended to offend Buddhism when he posted the image, and \\ after realizing it had caused widespread outrage, deleted it and issued an apology.\end{tabular} & 0.82 \\
MAP3 & \begin{tabular}[c]{@{}l@{}}At a hearing in December, Blackwood said he had not intended to offend Buddhism when he posted the image, and \\ after realizing \textbf{realizing} it had caused widespread outrage, deleted it and issued an apology.\end{tabular} & 0.81 \\
MAP4 & \begin{tabular}[c]{@{}l@{}}At a hearing in December, Blackwood said he had not intended to offend Buddhism \textbf{party} when he posted the image, and \\ after realizing it had caused widespread outrage, deleted it and issued an apology.\end{tabular} & 0.82 \\
MAP5 & \begin{tabular}[c]{@{}l@{}}At a hearing in December, Blackwood said he had not intended to offend Buddhism when he posted the image, and \\ after realizing it had caused widespread \textbf{Ferris}, deleted it and issued an apology.\end{tabular} & 0.80 \\
MAP6 & \begin{tabular}[c]{@{}l@{}}\textbf{at} a hearing in \textbf{japan}, \textbf{bailey} \textbf{admitted graham did} not intended to offend \textbf{buddhism} when\textbf{ buddhist} posted the \textbf{video}, and \\ after realizing he has caused widespread outrage, deleted it and issued \textbf{her} apology.\end{tabular} & 0.77 \\
MAP7 & \begin{tabular}[c]{@{}l@{}}At a hearing in December, Blackwood said he \textbf{lack} not intended to \textbf{keep} Buddhism when he posted the image, and \\ after realizing it \textbf{refuse} caused widespread outrage, \textbf{record} it and \textbf{recall} an apology.\end{tabular} & 0.76 \\
\begin{tabular}[c]{@{}c@{}}MAP8\\ (Russian)\end{tabular} & \begin{tabular}[c]{@{}l@{}}\foreignlanguage{russian}{На слушании в декабре Блэквуд сказал, что не имел намерения оскорбить буддизм, когда размещал изображение, а} \\ \foreignlanguage{russian}{после того, как осознал, что оно вызвало массовое возмущение, удалил его и опубликовал извинение.}\end{tabular} & 0.83 \\ \bottomrule
\end{tabular}%
}
\caption{An example of each MAP from our dataset for Ru-En. `Translation'  is the original machine translated sentence for the given source sentence, which was assigned an average DA score of 0.88 by human annotators (in 0-1). S1 are scores from MonoTransQuest architecture, and the reference translation is only shown for readability, as it was not used by humans nor QE models.}
\label{tab:mapexamples}
\end{table*}

\section{QE Models}
\label{sec:qemodels}

We choose three categories of heavy- to light-weight models for sentence-level QE models: first, the SOTA TransQuest with three variants MonoTransQuest, SiameseTransQuest and MultilingualTransQuest~\citep{ranasinghe-etal-2020-transquest-translation}; second, the LSTM-based Predictor-Estimator approach~\citep{kim-etal-2017-predictor} and third, the unsupervised method SentSim~\citep{song-etal-2021-sentsim}.

\paragraph{MonoTransQuest (MonoTQ)} This regression architecture encodes a concatenated source-target sentence pair using a transformer encoder. The architecture adds a softmax layer on top of the CLS token of the transformer to predict the quality of the translation. MonTransQuest architecture has separate pretrained QE models based on XLM-Roberta-Large~\citep{conneau-etal-2020-unsupervised} for all seven language pairs from WMT 2020 QE Task 1.

\paragraph{SiameseTransQuest (SiameseTQ)} This architecture uses a siamese network with two transformer models to encode the source and the target sentences separately. The architecture adds a max-pooling layer on top of the token embeddings of each transformer and calculates the cosine similarity between the outputs of the two pooling layers to predict the quality of the translation. Similar to \textit{MonoTQ}, SiameseTQ has separate pretrained QE models based on XLM-Roberta-Large for all seven language pairs from WMT 2020 QE Task 1.

\paragraph{MultilingualTransQuest (MultiTQ)} This architecture is based on MonoTQ but is trained on aggregated QE data for all seven language pairs from the WMT 2020 QE Task 1, resulting in one model for all the language pairs. This model is also based on XLM-Roberta-Large. 

\paragraph{Predictor-Estimator (OpenKiwi)} This is a two-stage architecture, where the Predictor model is an encoder-decoder RNN trained on parallel data (source-reference); in this case, the same data is used to train the respective NMT model for each language pair. Its output is then fed to the Estimator, a unidirectional RNN trained on QE data, to produce the quality estimates. Compared to TransQuest, the PredEst architecture does not rely on heavily pre-trained representations, resulting in a lighter model. For our experiments, we use the implementation in OpenKiwi~\citep{kepler-etal-2019-openkiwi}, which was provided as the baseline for the WMT 2020 QE Shared Task.

\paragraph{SentSim} This is an unsupervised method to QE that uses a combination of cross-lingual word and cross-lingual sentence similarity scores to produce a sentence-level quality score. The word-level similarity is extracted using BERTScore~\citep{zhang2019bertscore} between source and MT sentences, while sentence-level similarity is measured as the cosine similarity between the source and MT sentences representations. Both word and sentence-level representations are extracted using a cross-lingual pretrained model, namely, XLM-Roberta-Base~\citep{conneau-etal-2020-unsupervised}. 

\section{Probing Strategies}\label{sec:probing}

In this section, we introduce the rationale for two types of probes:  meaning-preserving and meaning-altering perturbations. We then describe each perturbation and discuss the experimental setup for the probing.

We define a {\bf meaning-preserving perturbation (MPP)} as a small change in the target-side translation that might affect the translation but should not affect the overall meaning of the sentence. For example, removing punctuation marks from the translated sentence should not affect the meaning conveyed by the text. By contrast, {\bf meaning-altering perturbations (MAP)}  should alter the meaning conveyed by the translation, for example, replacing a random word with its antonym or randomly replacing a content word. By introducing MAPs, we focus on probing models for whether they capture (lack of) adequacy in translations. Given that SOTA QE models are based on pre-trained representations obtained from strong language models, it has been hypothesised that they could be biased by the fluency of translations~\citep{sun-etal-2020-estimating}. 

We, therefore, design two types of perturbations: MPPs, which might affect fluency but not adequacy, and MAPs, which affect adequacy. Perturbations are only introduced in the translations to mimic translation errors. We have chosen perturbations that can be introduced using automated methods, and we carefully select perturbations relevant for MT, \textit{e.g.,} rare errors such as the omission of negation, and known errors such as omission of words from translation.  Each type of perturbation is introduced independently of others, one perturbation per target sentence.  We note that most of our perturbations are general enough such that they apply to all sentences in our test set. An exception is the removal of negation which can only be applied to sentences which contain a negation marker.

We analyse the behaviour of QE models by comparing the difference in the scores predicted after MPP/MAPs are applied to the test set compared to the original, unperturbed test set. We expect a strong QE model to predict lower scores to the version of the test set containing sentences with MPP and MAP, and -- more importantly, a higher score to sentences with MPP than to sentences with MAP. Each of our probes is detailed below, categorised either as an MPP or as an MAP.

\subsection{Meaning-Preserving Perturbations}
\label{subsec:mpps}
We designed the following MPPs. In order to ensure sufficient randomisation of the experiments, we repeat MPP2, 4, 5 and 6, twenty times for each sentence and average the QE scores obtained for these twenty perturbations. Other MPPs, \textit{e.g.,} removing all punctuations in the translation, can only result in one new version of the translation, and therefore, repetitions are not needed. 
\begin{description}
    \item[Removal of Punctuations (MPP1):] We remove any punctuation marks from the translation using the standard \textit{string} library in Python, for this perturbation. 
    \item[Replacing Punctuations (MPP2):] In this perturbation, each punctuation mark in the translation is replaced with another randomly chosen punctuation mark.
    \item[Removal of Determiners (MPP3):] We use the  \textit{spaCy\footnote{\href{https://spacy.io/}{spaCy API}}} Part-of-speech (POS) tagger to identify determiners, and then remove them from the translation.
    \item[Replacing Determiners (MPP4):] Each word labelled as a determiner with the help of spaCy POS tagger in the translation is replaced with another randomly chosen determiner from a list.
    \item[Change in Word-casing (MPP5/MPP6):] We select random content words from the translation and convert them to UPPERCASE to generate a set of perturbed translations (MPP5). Additionally, we select content words randomly from the translation and convert them to lowercase to generate another set of perturbed translations (MPP6). 
\end{description}

For each of the perturbations described above, we provide an example in Table~\ref{tab:mppexamples}, along with the scores predicted from our SOTA (MonoTQ) QE system.

\subsection{Meaning-Altering Perturbations}
\label{subsec:map}

We choose the following probes as MAP. We ensure sufficient randomisation of the experiments by repeating MAP2, 3, 4, 5, 6, and 7, twenty times for each sentence, and average the QE scores obtained for these twenty perturbations. For MAPs 1, and 8, we can produce only one version of the sentence.  

\begin{description}
    \item[Removal of Negation Markers (MAP1):] For this perturbation, all the \textit{negation markers} like ``no'', ``not'', ``n't'' \textit{etc.} are removed. 
    \item[Removal of Random Content Words (MAP2):] We select a random content word from the translation and remove it. 
    \item[Duplication of Random Content Words (MAP3):] We choose a random content word from the translation and add it at the immediate next position index, thus duplicating its occurrence. 
    \item[Insertion of Random Words (MAP4):] We populate a vocabulary of words from the complete set of translations in our test set. From this vocabulary, we choose a word and insert it at a random position in the sentence, ensuring that the previous word and the next word are not the same to avoid duplication.
    \item[Replacing Random Content Words (MAP5):] We choose a random content word from the translation and replace it with another word from the vocabulary created as discussed in MAP4. 
    \item[BERT-based Sentence Replacement (MAP6):] We obtain sentence replacements based on the BERT-base model~\citep{devlin-etal-2019-bert}, with the help of a data augmentation library\footnote{\label{note1}\href{https://github.com/makcedward/nlpaug}{GitHub: makcedward/nlpaug}}~\citep{ma2019nlpaug}. This library uses a word replacement approach proposed by~\citet{kobayashi2018contextual} and generates a sentence synonymous to the input provided. We observe that BERT-generated synonymous sentences replace content words which alter the inherent meaning of the input sentence and hence, treat this perturbation as MAP.
    \item[Replacing Words with Antonyms (MAP7):] With the help of the data augmentation library\footnoteref{note1}, we generate perturbed translations where we replace random words in the sentence with their antonyms from the English Wordnet~\citep{miller1990introduction}.
    \item[Source Sentence as Target (MAP8):] We replace the translation with the source side sentence to observe the effect on QE scores when the source sentence is evaluated by the QE model, instead of the target side translation. Such a perturbation results in the model input to become \textit{source-source} instead of \textit{source-target}.
\end{description}

For each of the perturbations described above, we provide an example in Table~\ref{tab:mapexamples}, along with the scores predicted via SOTA (MonoTQ) system.

\begin{figure*}[ht!]
    \centering
    \includegraphics[width=1\textwidth, height=22cm]{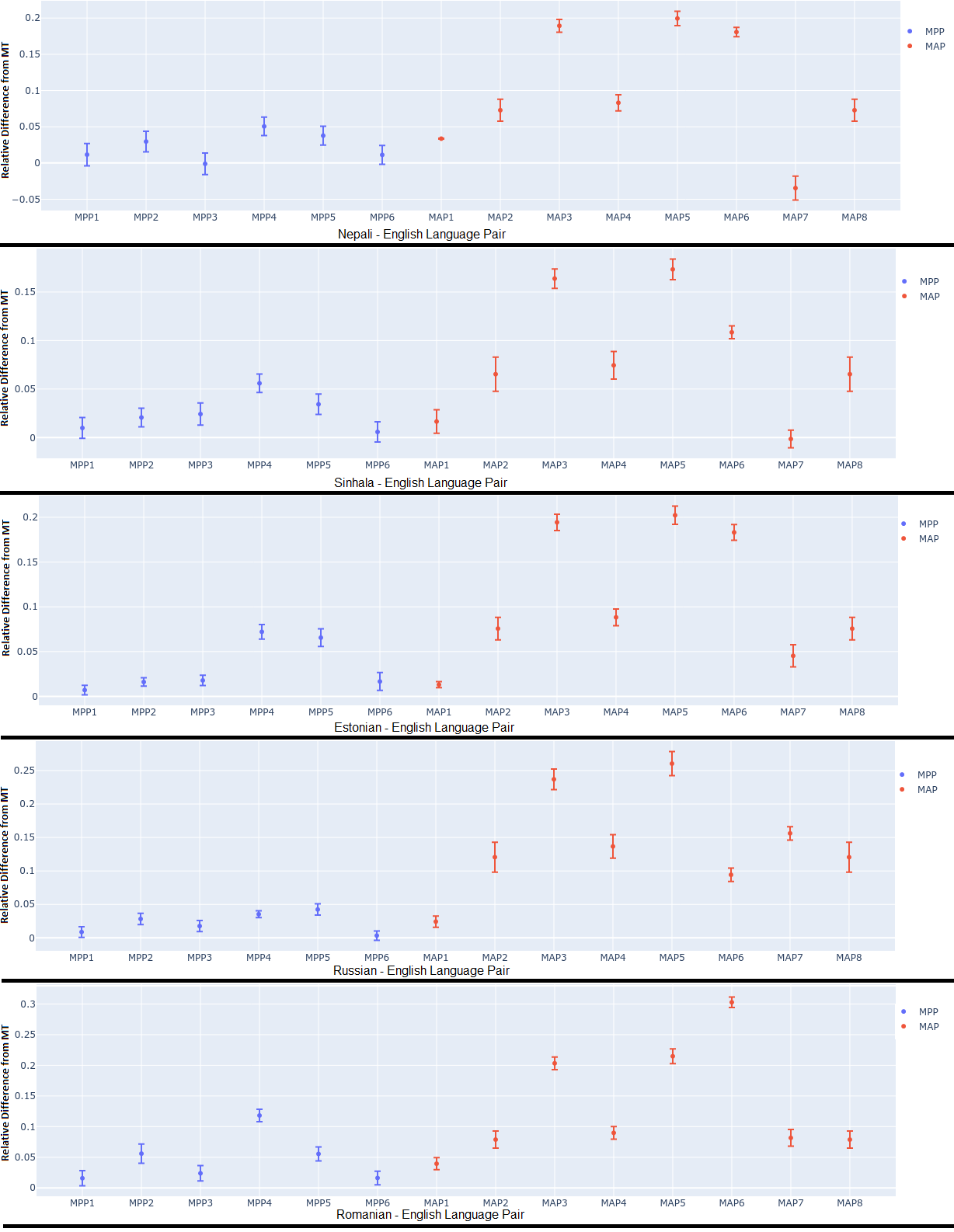}
    \caption{Average difference between the predicted QE scores for original translations and each perturbation across the test set for each language pairs (Y-axis->MT - $x$, where $x$ is the perturbation as labelled on the X-axis), using the SOTA MonoTQ architecture.}
    \vspace{-0.2cm}
    \label{fig:graphone}
\end{figure*}


\begin{table*}[t!]
\centering
\resizebox{\textwidth}{!}{%
\begin{tabular}{@{}cccc|ccc|ccc|ccc|ccc@{}}
\toprule
 & \multicolumn{3}{c|}{Ru-En} & \multicolumn{3}{c|}{Ro-En} & \multicolumn{3}{c|}{Et-En} & \multicolumn{3}{c|}{Si-En} & \multicolumn{3}{c}{Ne-En} \\ \midrule
 & MT & MPP & MAP & MT & MPP & MAP & MT & MPP & MAP & MT & MPP & MAP & MT & MPP & MAP \\
MonoTQ & 0.81 & 0.78 & \textbf{0.66} & 0.82 & 0.80 & \textbf{0.74} & 0.81 & 0.79 & \textbf{0.73} & 0.71 & 0.65 & \textbf{0.64} & 0.75 & 0.74 & \textbf{0.68} \\
SiameseTQ & 0.86 & \textbf{0.85} & 0.86 & 0.58 & 0.57 & \textbf{0.52} & 0.92 & \textbf{0.91} & \textbf{0.91} & 0.58 & 0.57 & \textbf{0.52} & 0.68 & 0.68 & \textbf{0.65} \\
MultiTQ & 0.79 & 0.75 & \textbf{0.68} & 0.79 & 0.74 & \textbf{0.66} & 0.77 & 0.73 & \textbf{0.66} & 0.62 & 0.58 & \textbf{0.52} & 0.63 & 0.60 & \textbf{0.52} \\
OpenKiwi & 0.78 & 0.78 & 0.78 & 0.78 & \textbf{0.75} & 0.77 & 0.71 & \textbf{0.70} & \textbf{0.70} & 0.62 & 0.60 & \textbf{0.57} & 0.50 & \textbf{0.48} & \textbf{0.48} \\
SentSim & 0.54 & 0.57 & 0.57 & 0.78 & 0.76 & \textbf{0.72} & 0.50 & 0.53 & 0.52 & 0.41 & 0.43 & 0.41 & 0.47 & 0.52 & 0.50 \\ \bottomrule
\end{tabular}%
}
\caption{Average predicted scores by all QE models on the test set for the original (unperturbed) machine translation (MT), versus its version with meaning-preserving perturbations (MPP) and  meaning-altering perturbations (MAP). Between MPP and MAP, we boldface the lowest average scores, if lower than MT.}
\label{tab:resultsaveragescores}
\end{table*}

\section{Results and Discussion}
\label{sec:results}

In this section, we discuss the results obtained from our probing experiments using various QE models. 

\subsection{Do perturbations affect SOTA QE models?}

We start by analysing the behaviour of MonoTQ, as the best performing SOTA QE model on the dataset used in this paper, under different types of perturbations. Figure~\ref{fig:graphone} shows the difference between the average predicted score for our original test set (Table~\ref{tab:dataset}) before perturbations and the same subset of sentences perturbed using MPP and MAP. In comparison to the average scores for the initial set of translations, the expected behaviour for a strong QE model is to assign the same or slightly lower scores to and their MPP counterparts, but substantially lower scores to the MAP variants. Based on this premise, we can make the following observations from Figure~\ref{fig:graphone}. The other graphs obtained from SiameseTQ model, MultiTQ model, OpenKiwi system, and the Unsupervised method are present in Appendix \ref{sec:appendix}.

\paragraph{Models are robust to MPPs and sensitive to MAPs} Overall, sentences with MPPs result in a small drop in the scores with respect to the original set of translations, especially when compared to the sentences containing MAPs. Conversely, perturbations that affect sentence meaning have a larger impact on the scores. Thus, SOTA QE models are indeed capable of discriminating between the two types of changes. 

\paragraph{Models fail to detect important MAPs}
However, MonoTQ fails to discriminate between MPPs and specific types of MAPs. In particular, \textit{perturbations that affect sentence polarity}, \textit{i.e.,} MAP1 (Removal of Negation Markers) and MAP7 (Replacing Words with Antonyms) result in a similar drop in the predicted scores as MPPs. An exception is a slight increase in the case of Nepali-English where the number of instances with negation markers were limited to only 4, which makes it impossible to draw any conclusions. Omitting negation is a critical error in the practical applications of MT. But it does not frequently occur in the data, and therefore, cannot be detected by using the standard way of assessing the performance of QE systems, \textit{i.e.,} by computing the correlation with human judgements on a test set.

MAPs that correspond to \textit{omission and addition errors in translation} (MAP2 and MAP4, respectively) also result in a relatively small drop in the predicted scores and thus hardly be distinguished from MPPs. Omitting contents is a well-known issue for the current neural MT models~\citep{yang-etal-2019-reducing}. An omission is particularly dangerous as it can go unnoticed by the end-user of the MT system. The ability to detect such errors is thus a crucial task for QE and, as highlighted by our analysis, requires further work in this direction.

Finally, \textit{copying the source sentence in the translation} (MAP8) is not adequately captured by MonoTQ. Note that this represents another critical translation error, as the source sentence is left untranslated. We hypothesise that the inability to detect copy errors is due to the fact that MonoTQ relies on the multilingual pre-trained representations and, unless presented with such cases during fine-tuning, would treat the two sentences in the source language as equivalent.

\paragraph{Comparison across languages} Interestingly, we observe similar trends across language pairs. For all the language pairs, sentences with MAP produce a larger drop in performance than MPP, and the same MAPs result in incorrect behaviour.
\begin{figure*}[ht!]
    \centering
    \includegraphics[width=1\textwidth]{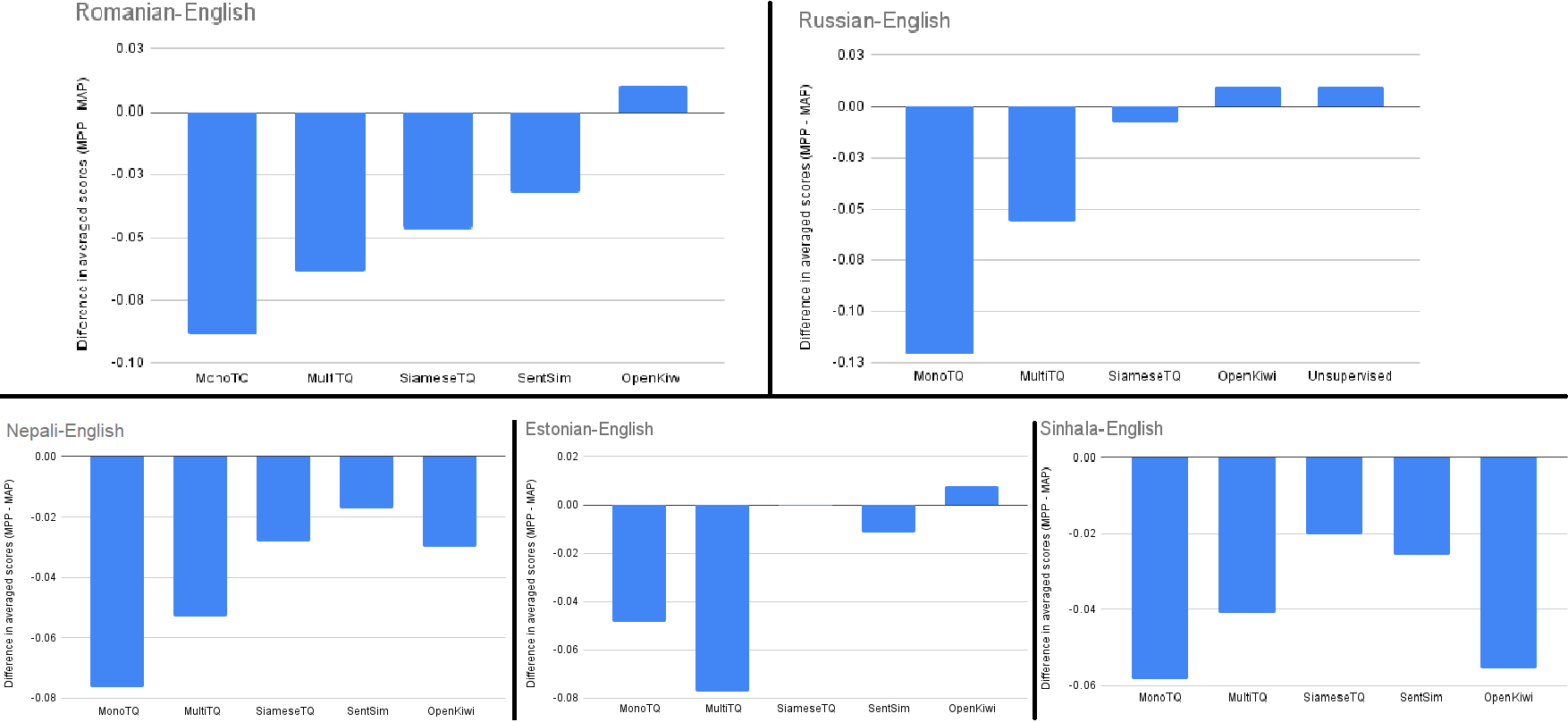}
    \caption{Ranking QE models using our method (MPP - MAP), where different QE models are shown on the X-axis, sorted as per the ranks obtained via Pearson correlation (among QE scores and human DA judgements). The size of the bars corresponds to the ability of the QE models to distinguish between MAP and MPP perturbations - the higher the negative bar, the better the QE model is at this task.}
    \label{fig:rankfigall}
\end{figure*}
\subsection{Do perturbations affect other QE models?}

Table~\ref{tab:resultsaveragescores} shows the actual average scores produced by different QE systems for the initial subset of high-quality MT sentences (column MT) and the same subset of sentences perturbed using MPP (column MPP) and using (column MAP). For strong QE models, we would expect both MPP and MAP scores to be lower than the initial MT outputs, especially for MAP. For most of the models and languages, the sentences perturbed with MAP receive lower average scores, thus confirming that, in general, QE models are sensitive to the changes that affect meaning. It is clear, however, that for some models, the difference between the  MT, MAP and MPP is negligible. These cases are observed with OpenKiwi and SentSim, which are weaker QE models compared to the TransQuest variants~\citep{specia-etal-2020-findings-wmt} (see Table~\ref{tab:pearsonall} for the overall results on the complete test+validation set of 2K sentences). Thus, we hypothesise that the ability of a QE model to discriminate between MAP and MPP could be predictive of its overall performance. We empirically test this hypothesis, and discuss below.

\subsection{Can we use perturbations to rank QE models?}

We pose that the overall performance of a QE system can be predicted based on how well it is able to discriminate between meaning-preserving and meaning-altering perturbations. To test this claim, we contrast the ability of a set of QE systems to discriminate between MAP and MPP with their overall performance measured in terms of Pearson correlation with human judgements.  Table~\ref{tab:pearsonall} shows sentence-level Pearson correlation with human judgements on the WMT 2020 QE Shared Task test set for all the QE models and language pairs considered in our experiments. As shown in Table~\ref{tab:pearsonall}, QE models vary a lot in terms of overall performance, the weakest system being OpenKiwi and SentSim, and the strongest corresponding to the SOTA approaches based on XLM-Roberta. To assess the discriminative power of the models, we compute the average difference (MPP - MAP) between the relative scores obtained via our method (such as shown in Figure~\ref{fig:graphone}). In Figure~\ref{fig:rankfigall}, we sort all the probed QE models in the decreasing order, according to the correlation with human judgements on the x-axis, and plot the corresponding MAP/MPP difference on the y-axis.

\begin{table}[ht!]
\centering
\resizebox{\columnwidth}{!}{%
\vspace{-0.5cm}
\begin{tabular}{@{}cccccc@{}}
\toprule
 & \textbf{Et-En} & \textbf{Ru-En} & \textbf{Ro-En} & \textbf{Si-En} & \textbf{Ne-En} \\ \midrule
\textbf{MonoTQ} & 0.72 & \textbf{0.77} & \textbf{0.88} & \textbf{0.88} & \textbf{0.75} \\
\textbf{MultiTQ} & \textbf{0.76} & \textbf{0.77} & 0.87 & 0.87 & 0.74 \\
\textbf{SiameseTQ} & 0.55 & 0.71 & 0.84 & 0.84 & 0.60 \\
\textbf{SentSim} & 0.53 & 0.46 & 0.77 & 0.77 & 0.56 \\
\textbf{OpenKiwi} & 0.47 & 0.59 & 0.68 & 0.36 & 0.39 \\ \bottomrule
\end{tabular}%
}
\caption{Pearson correlation with human judgements for all QE models on the original, complete test+validation (2K) set. This is the metric used to rank participating QE systems in the WMT 2020 QE Shared Task 1. As can be seen, MonoTQ and MultiTQ consistently outperform all other models, with OpenKiwi performing the poorest.}
\vspace{-0.5cm}
\label{tab:pearsonall}
\end{table}

Interestingly, for most of the language pairs, we observe that the system rankings are similar or identical to the Pearson correlation-based rankings; indicating that the ability of the model to distinguish between the proposed types of perturbations is indeed indicative of its overall performance. One exception is the difference corresponding to the OpenKiwi system for Sinhala-English and Nepali-English. We attribute this to the fact that, by difference from the SOTA QE models, OpenKiwi is good at capturing the copy errors (MAP8) for these languages. OpenKiwi uses different vocabularies for the source and target languages, and therefore, copying the source sentence results in unknown tokens on the target side, leading to a low predicted score. Another exception is Estonian-English, where the systems appear to be ranked differently based on correlation vs. MAP/MPP difference. We note, however, that even in this case, the two top-performing systems (MonoTQ and MultiTQ) are clearly distinguished from the low-performing ones (SentSim and OpenKiwi).

Although generating MAPs and MPPs requires some initial set of high-quality translations, this could be selected using reference sentences from parallel data. Therefore, the proposed methodology allows for assessing the performance of QE models with no need for collecting explicit human judgements (\textit{e.g.,} direct assessments).

\section{Conclusions}

In this work, we have proposed a methodology for analysing the performance of QE systems beyond correlation with human judgements. We have devised a set of perturbations to probe both the robustness of QE models towards changes in the input that do not affect sentence meaning and their sensitivity to meaning errors in translation. First, by applying the proposed methodology to a set of QE systems of varying accuracy, we are able to detect specific failures that cannot be detected by computing correlations between predicted scores and human judgements. Second, we have shown that, on an average, the ability of a given model to discriminate between the two types of perturbations is predictive of its overall performance, thus allowing us to compare QE systems without relying on manual quality annotation.

Our choice of specific perturbations was motivated by the errors that occur in neural MT and the potential weaknesses of QE models. In the future, we plan to extend this set by including perturbations that capture other critical MT errors. Furthermore, we plan to study whether the proposed perturbations can be used at training time to improve the ability of QE systems to detect critical errors in translation.

\bibliography{anthology,custom}
\bibliographystyle{acl_natbib}
\clearpage

\appendix
\section{Appendix}
\label{sec:appendix}

\begin{minipage}{\textwidth}
\begin{minipage}{\textwidth}
    \centering
    \includegraphics[width=0.9\textwidth]{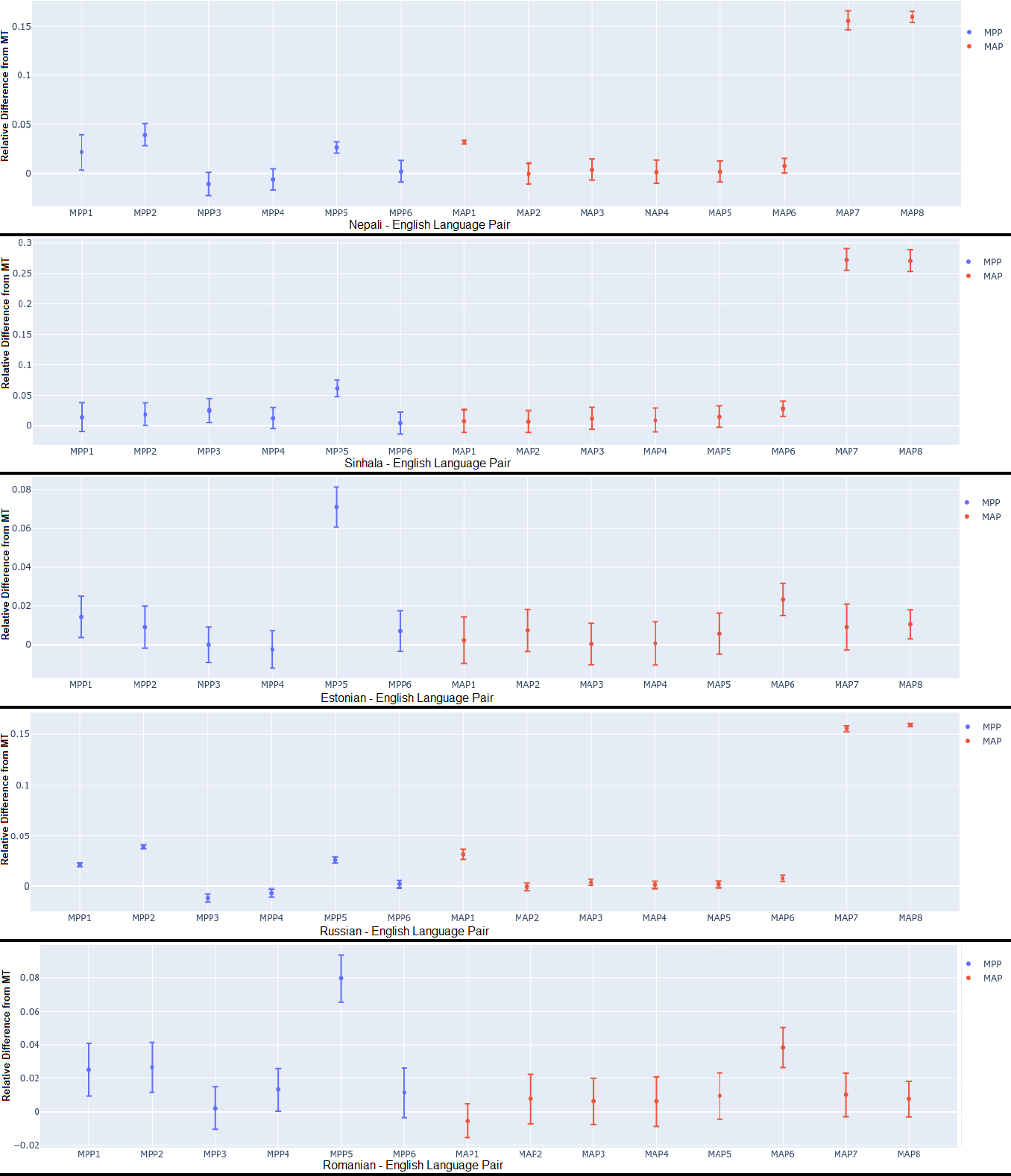}
    \captionof{figure}{Difference between the predicted QE scores for original sentences and each perturbation for all language pairs (Y-axis->MT - $x$, where $x$ is the perturbation as labelled on the X-axis), using the \textbf{\em OpenKiwi} system.}
    \label{fig:openkiwiall}
\end{minipage}
\end{minipage}

\begin{figure*}[ht!]
    \centering
    \includegraphics[width=\textwidth]{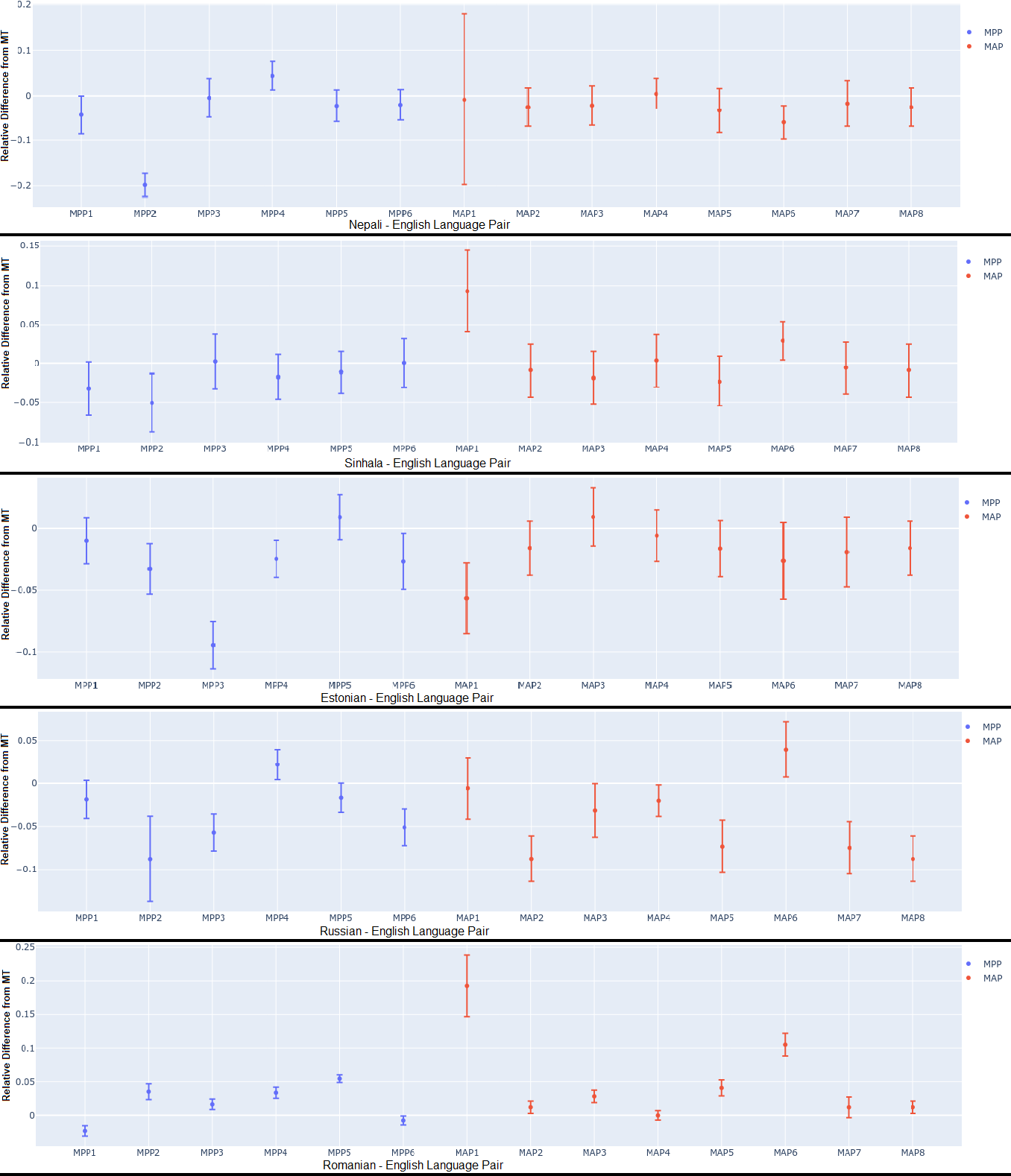}
    \caption{Difference between the predicted QE scores for original sentences and each perturbation for all language pairs (Y-axis->MT - $x$, where $x$ is the perturbation as labelled on the X-axis), using \textit{\textbf{Unsupervised \textit{SentSim}}} method.}
    \label{fig:sentsimall}
\end{figure*}
%
\begin{figure*}[ht!]
    \centering
    \includegraphics[width=\textwidth]{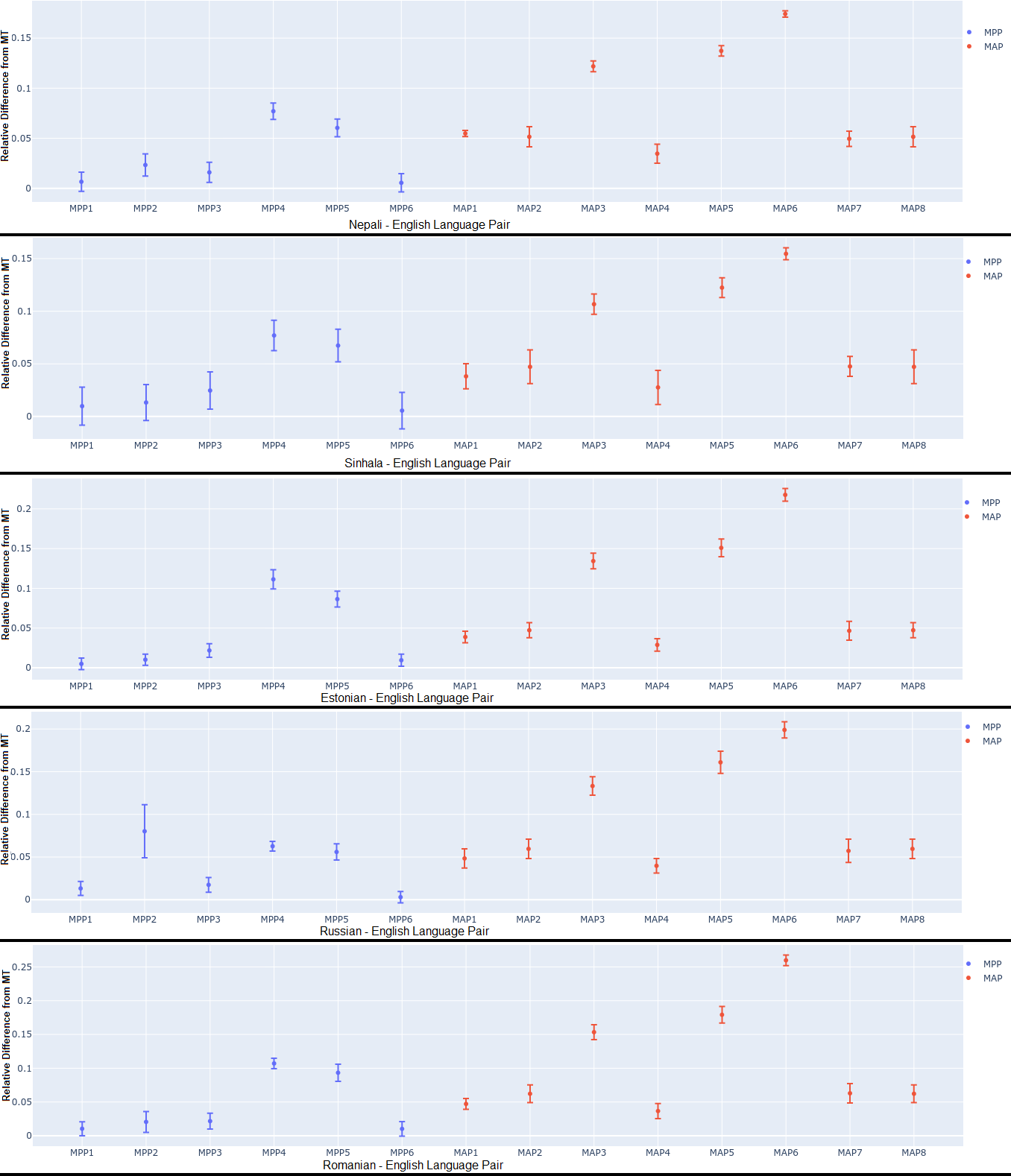}
    \caption{Difference between the predicted QE scores for original sentences and each perturbation for all language pairs (Y-axis->MT - $x$, where $x$ is the perturbation as labelled on the X-axis), using \textit{\textbf{MultilingualTransQuest}} architecture.}
    \label{fig:multitqall}
\end{figure*}
\begin{figure*}[ht!]
    \centering
    \includegraphics[width=\textwidth]{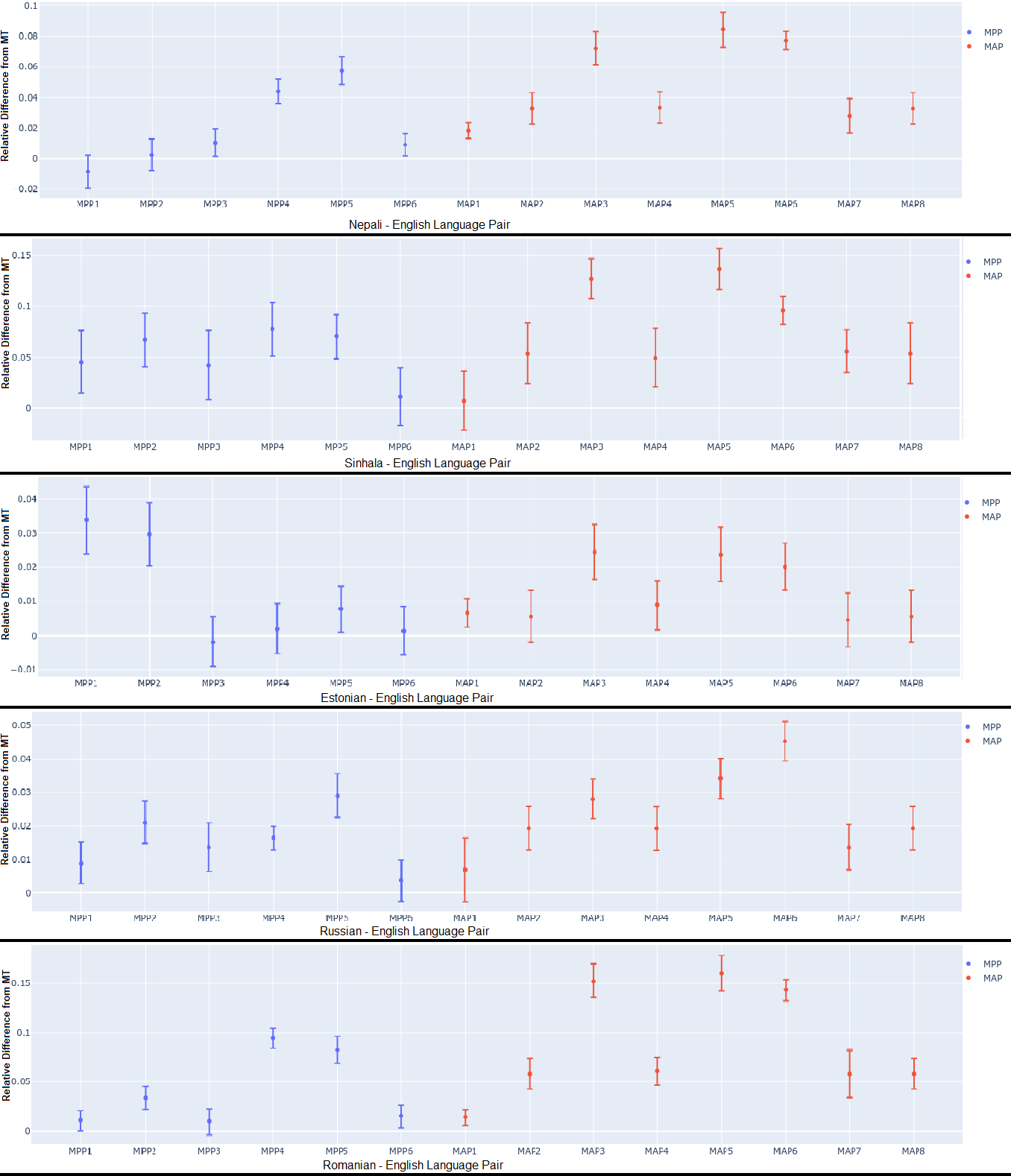}
    \caption{Difference between the predicted QE scores for original sentences and each perturbation for all language pairs (Y-axis->MT - $x$, where $x$ is the perturbation on the X-axis), using the \textit{\textbf{SiameseTransQuest}} architecture.}
    \label{fig:siamesetqall}
\end{figure*}
\end{document}